# Detecting Cross-Lingual Plagiarism Using Simulated Word Embeddings

Victor U Thompson and Chris Bowerman

Department of Computer Science

University of Sunderland, SR1 3SD

*Abstract*: Cross-lingual plagiarism (CLP) occurs when texts written in one language are translated into a different language and used without acknowledging the original sources. One of the most common methods for detecting CLP requires online machine translators (such as Google or Microsoft translate) which are not always available, and given that plagiarism detection typically involves large document comparison, the amount of translations required would overwhelm an online machine translator, especially when detecting plagiarism over the web. In addition, when translated texts are replaced with their synonyms, using online machine translators to detect CLP would result in poor performance. This paper addresses the problem of cross-lingual plagiarism detection (CLPD) by proposing a model that uses simulated word embeddings to reproduce the predictions of an online machine translator (Google translate) when detecting CLP. The simulated embeddings comprise of translated words in different languages mapped in a common space, and replicated to increase the prediction probability of retrieving the translations of a word (and their synonyms) from the model. Unlike most existing models, the proposed model does not require parallel corpora, and accommodates multiple languages (multi-lingual). We demonstrated the effectiveness of the proposed model in detecting CLP in standard datasets that contain CLP cases, and evaluated its performance against a state-of-the-art baseline that relies on online machine translator (T+MA model). Evaluation results revealed that the proposed model is not only effective in detecting CLP, it outperformed the baseline. The results indicate that CLP could be detected with state-of-the-art performances by leveraging the prediction accuracy of an internet translator with word embeddings, without relying on internet translators.

## 1   Introduction

Cross-lingual plagiarism (CLP) occurs when texts written in one language are translated into another language and used without acknowledging the original sources. Extensive studies have been carried out on monolingual plagiarism analysis which involves searching for plagiarism in documents of the same language, but CLP still remains a challenge. Access to information via the internet and the availability of online translation tools (such as Google and Microsoft translation services) make it easy to acquire and translate information from one language to another, and use them without being detected. The obvious challenge in CLPD is the disambiguation of different languages. Previous studies have addressed this problem using methods such as statistical machine translation (Barrón-

Cedeño et al., 2013), cross-lingual explicit semantic analysis (CL-ESA) (Potthast et al., 2011), syntactic alignment using character N-grams (CL-CNG), dictionaries and thesaurus (Pataki, 2012, Gupta et al., 2012), online machine translators (Ehsan et al., 2016; Ferrero et al., 2017c), and more recently, semantic networks and word embeddings (Franco-Salvador et al., 2014, 2016; Speer. and Lowry-Duda, 2017; Ferrero et al., 2017a; Glavaš et al., 2017). Most of the proposed models are either limited to bilingual CLPD tasks, require parallel or comparable corpus which are usually not sufficient or available for low resource languages, while others rely on internet translation services, which are not available for large scale CLPD.

The most common and intuitive approach for detecting CLP involves using internet translation services to normalise texts written in different languages into a common language before applying monolingual plagiarism analysis to detect plagiarism. This approach is known as the translation plus monolingual analysis (T+MA), and has been exceptional in previous studies when compared with other state-of-the art CLPD models (Barrón-Cedeño et al., 2013). However, one major drawback of the T+MA model is that internet translation services are not always available, and the amount of translations required for detecting plagiarism in large collections such as the internet is overwhelming. Hence as correctly pointed out in (Meuschke and Gipp 2013), CLPD methods that rely on internet translation services are only suitable when working with small document collections. In addition, there are severe cases where translated texts are further altered (paraphrased) by replacing words in texts with their synonyms to obfuscate plagiarism and evade detection; the T+MA model and most of the other state-of-the-art models are unlikely to be effective under such conditions (Franco-Salvador et al., 2014; 2016).

In spite of the limitations mentioned above, the T+MA is still the most common CLPD model. An evaluation of state-of-the-art CLPD models (Barrón-Cedeño et al., 2013) revealed that the T+MA outperformed the others, which is likely due to its high precision in detecting translated words, given that most cases of CLP are formed using online translation tools similar to what the T+MA uses (Google translate). Hence applying the T+MA model to detect such cases of CLP simply requires retranslating texts back to their original language using the same online machine translator. The question then is how to capture the precision and overcome the limitations of the T+MA model, without losing recall.

Recent advancement in word embeddings revealed efficient and effective embedding models such as the word2vect (Mikolov et al., 2013a) and Glove (Pennington et al., 2014), both of which are able to predict semantically similar words with high accuracy. These models are based on the intuition that words that appear frequently in the same contexts are considered similar, and have been recently extended to CL word similarity analysis by using a common representation space for more than one language (Ruder et al., 2017). Most applications of CL-WE in CLPD are designed for bilingual tasks

(for only two languages), and require parallel corpora (Ferrero et al., 2017a; Glavaš et al., 2017). In this work, we propose a CLPD model that uses CL-WE to capture the translations of words (in multiple languages) from a common online machine translator and apply the translated words in CLPD. The proposed model is designed for both bilingual and multilingual CLPD tasks, and does not require the use of internet translators or parallel corpora. Evaluation of the proposed CLPD model revealed that, it is not only effective in detecting CLP, it outperformed a standard T+MA baseline.

This paper makes the following contributions:

1) We propose a CLPD model that leverages word embeddings to detect CLP cases that contain texts that are directly translated to their semantic equivalent in other languages, and cases where translated words are replaced with related words (synonyms). The proposed model can be used for multilingual CLPD, and not just for bilingual tasks; it is language independent and does not require comparable or parallel corpora.
2) We propose an efficient and effective method for creating bilingual and multilingual texts (corpora) for building CLPD models for low resource languages that have little or no parallel or comparable corpora.
3) We propose a method that captures online machine translations across multiple languages and apply them in CLPD without relying on the internet.

The rest of this paper includes related work on CLP, research question, the proposed CLPD and multilingual translation models, evaluation of the proposed CLPD model, results and implications, conclusion and future work.

## 2   Related Studies

This section describes state-of-the-art methods used in detecting CLP, some of which have been in existence for a while relative to newer models. Existing models include cross-lingual character n-grams (CL-CNGs) that uses character n-grams to measure the syntactic similarity between texts, cross-lingual explicit semantic analysis (CL-ESA) (Potthast et al., 2011) that uses comparable (intermediary) corpora to capture topic similarity, cross-lingual alignment-based similarity analysis (CL-ASA) (Barrón-Cedeño et al., 2013) that uses statistical machine translation to align parallel corpora, thesaurus (Gupta et al., (2012) and dictionary (Pataki, 2012) methods that rely on external resources such as Eurovoc and multi-lingual dictionaries to measure CL similarity, and finally the T+MA model that normalises texts into a common language using an online machine translator, before applying monolingual plagiarism detection methods. Of these models, the T+ MA is the most common, having been used by most participants in the Pan competition on plagiarism detection, and in the SemEval (Agirre et al., 2016; Cer et al., 2017) competition on cross and multi lingual semantic similarity analysis, including the best performing systems (Tian et al., 2017; Wu et al., 2017).

However, the T+MA model is limited by the fact that internet translation tools are not always available, and it is not effective in situations where texts are translated and then altered by replacing words with similar words. The T+MA model is also limited by the amount of translations required for large scale plagiarism detection over the internet.

**Newer Approaches for CLPD**: In more recent studies, approaches based on word embeddings and semantic networks have been proposed for CLPD. Word embedding models use distributed representation of words to predict semantically similar words; the basic idea is that words that appear frequently in the same contexts are considered similar. Common but efficient (and effective) word embedding architectures include the wor2vec CBOW and skip-gram models (Mikolov et al., 2013a), and Glove (Pennington et al., 2014); (Ghannay et al., 2016). These models map words to vectors of real numbers, and follow the intuition that when words are represented in a common vector space, similar words should have similar vector representations. Word embeddings were originally proposed for monolingual similarity analysis, but have recently been extended to cross-lingual settings where a joint embedding space is used to learn cross-lingual representation of words (Ruder et al., 2017); typical implementation involves projecting the embeddings of a source language into the space of a target language. Common CL-WE models are the canonical correlation analysis (CCA) (Faruqui and Dyer, 2014; Lu et al., 2015; Ammar et al., 2016), alignment projection (Guo et al., 2015; 2016), and the linear transformation model proposed by Mikolov et al., (2013b). The CCA projects matrices from parallel corpora into lower dimensions of maximum correlation, and translate by projecting across both matrices. The alignment projection method aligns a parallel corpora (bilingual dictionary) and project words in a target language to the embedding space of a trained monolingual source. The linear projection method uses a linear transformation function to map the embeddings of one language into the space of another using a trained dictionary that learns the function. Duong et al., (2017) argued that most of the common CL-WE models were built for bilingual analysis, and proposed a method based on multi-lingual joint training that combines bilingual word embeddings from multiple languages in a common embedding space and obtained encouraging results. CL-WE are rarely used in CLPD, although they can be effective offline machine translators. In one application of CL-WE in CLPD, Ferrero et al., (2017a) used bilingual embeddings (from a parallel corpus) mapped to a common space, and trained using the word2vec CBOW model to detect CLP. The actual implementation was carried out using *MultiVec* (Berard et al., 2016); a toolkit designed for creating and managing a number of distributed representation models, and the specific model used was original proposed by Luong et al., (2015); a skip-gram extension of word2vec in a bilingual space. CLPD was carried out by comparing word vectors in pairs of suspect and source sentences using the cosine measure, pairs of sentences with similarity scores above a predefined threshold are considered potential plagiarised fragments. The skip-gram and CBOW models retain word order and capture the context of a word, which reveal a lot about the word. In a similar study Glavaš et al., (2017) applied the linear transformation method proposed by Mikolov et al., (2013b) a in CLPD; the actual detection of plagiarism was based on word vector comparison on sentence level similar to Ferrero et al., (2017a).

Semantic networks link words with similar meanings in different languages to common concepts; words more closely linked to a concept are assigned higher weights than words further away. Examples of semantic networks are BarbleNet and ConceptNet (Franco-Salvador et al., 2014, 2016; Speer. and Lowry-Duda, 2017). Franco-Salvador et al., (2014; 2016) proposed a CLPD model that uses knowledge graphs built form a multi-lingual semantic network (MSN) to compare documents in different languages for semantic similarity, and CLPD. Knowledge graphs capture relationships between concepts derived from textual fragments in a document, while a MSN links semantically related words in different languages to a specific concept. The knowledge graph method was proposed because most of the existing CLPD models were designed to detect texts that have been translated to other languages using online machine translators, and struggle to detect translated texts that have been paraphrased. To address the problem, Franco-Salvador et al., applied BableNet, a MSN made form concepts derived from Wikipedia and WordNet, to build knowledge graphs for documents written in different languages and apply a similarity function to measure their similarity. Results from Franco-Salvador et al., study shows that the knowledge graph method outperformed state-of-the-art methods.

Hybrid models that combine word embeddings with other methods have also been proposed. A hybrid of word embeddings and knowledge graph was proposed in (Franco-Salvado, 2016) with the aim of determining whether both models complement each other in detecting CLP, evaluation results presented by the authors revealed that the models are complementary, and that when in combination, they outperformed several state-of-the-art CLPD models. Speer and Lowry-Duda (2017) used concept-net to combine pre-trained word2vec and Glove models (word embedding models) into a multilingual semantic similarity detection system, and emerged best in the SemEval 2017 competition. Concept-net is an open multilingual knowledge graph that generates concepts that relate meanings of words and phrases. The idea used in combining the concept graph to a word embedding model is retrofitting, where a pre-trained embedding model is built upon a concept graph. This process is carried out separately on the individual word embedding models, and then combined using a unified vocabulary. The redundant features/dimensions that result from the combination are then removed via truncated singular value decomposing (SVD), a technique used in latent semantic analysis to reduce the dimensions of a VSM to only the most relevant ones. España-Bonet and Barrón-Cedeño (2017) presented a language independent model that measures the semantic similarity between text snippets across multiple languages. The system uses a Support Vector Machine (SVM) to combine a number of intertextual features, which includes features derived from embeddings trained using the word2vec model and a multi-lingual corpora, from lexical similarity measurements, from the internal representation (hidden layer) of a neural network trained using multi-lingual parallel corpora and from CL-ESA. This approach is however best suited for low resource languages.

Evaluation of state-of-the-art CLPD models (Barrón-Cedeño et al., 2013) revealed that the T-MA model outperformed the others due to its precision in translating texts using online translators, and as mentioned in (Burrows et al., 2013; Barrón-Cedeño et al., 2013), precision is the single most important measure used in plagiarism detection as it reduces the time in deciding whether plagiarism is carried out or not. Hence the objective of this work is to propose a CLPD model that leverages the precision of the T+MA model, while overcoming its limitations, with comparable performance.

# 3   Research Question

The question answered in this work is whether a CLPD model could be built to capture the precision and overall performance, but not the following limitations of a T+MA model:

- i)    Internet translation tools not always available,
- ii)   Overwhelmed by the amount of translations required for large scale CLPD over the web,
- iii)  Unable to detect CLP cases where texts are translated to a different language and then altered by replacing words with semantically related words (synonyms).

# 4   Methods

This section describes the proposed CLPD model and the methods used in the model. The CLPD model follows the standard architecture proposed in previous studies for detecting mono and cross lingual plagiarism (Potthast et al., 2011; Barrón-Cedeno et al., 2013), which include candidate selection, detailed comparison and extraction of plagiarised passages (text alignment), and post-processing. In the retrieval process, a suspect document is tokenised, keywords are extracted and expanded using a multi-lingual translation model, and the expanded query is used to retrieve candidate source documents from an inverted index built with a collection of source documents. Matched query words (during the candidate selection stage) and their corresponding source words are mapped to the sentences in which they appear in the suspect and candidate source documents respectively for detailed similarity analysis. Sentences with similarity scores above certain thresholds are used as plagiarised fragments to retrieve plagiarised passages. We begin by describing the proposed multi-lingual translation model (MTM) and how it was used in this work to detect CLP.

## 4.1   The Proposed Multi-Lingual Translation Model (MTM)

This subsection describes how the MTM is designed to capture the precision of a T+MA model. The MTM uses word embeddings to capture words and their translations in different languages; it is designed to reproduce the translation of words from an online machine translator when detecting CLP without using internet translators, and to detect semantically similar words (synonyms) by leveraging the potentials of a word2vec model in linking similar words that occur in different contexts in an embedding space. In summary, to build the MTM, generate the translations of words in different languages using Google translate (or any other online translation tool), map the words and their translations in a common space and replicate the embeddings to optimise performance, and then train the simulated embeddings using the Mikolov et al., (2013a) word2vec CBOW model.

The premise of the model is that similar words occur in similar contexts, if the probability of finding a word in a context is magnified, then higher similarity should be assigned to words that share similar contexts than to non-contextual words (words not in the context). The contexts are simulated, each comprises of a pivot word in English and its equivalent translations in other languages.

Example of a context: {*man (Eng), homme (French), mann (Germ), hombre (Spanish)*}

The context in the example above consists of the word '*man*' in English and its corresponding translations in French, German and Spanish. Our objective is to maximise the probability of retrieving a context *c* given a word *w*.

$$p(c \mid w) = \frac{\exp(c^T w)}{\sum_{c \in V} \exp(c^T w)}$$ , where *c* and *w* are the context and word embeddings, and *v* is the vocabulary. To create contexts for the model, we used the *top-k* most common English words (based on their frequency) from the British national corpus (BNC), and created context for each word by retrieving their translations in Danish, French, German, Spanish etc. from Google translate. The *top-K* could be the *top-10,000* word*s*, but *k* has to be carefully chosen for optimal performance, we experimented with different values of *k* to determine the best. Similar to Faruqui and Dyer (2014), we excluded the *top-100* most common words as being too common and non-discriminatory (noisy).

When contexts are created, the final stage is to train a feed-forward neural network, using the back propagation algorithm with stochastic gradient descent to learn the word distributions in the embeddings with the objective of maximising the conditional probability of retrieving an output word given an input context. However training the network this way would result in poor performance (inaccurate predictions) as sufficient data (statistics) is required about each context in order to increase the probability of retrieving a context when a word in the context is searched.

To achieve higher similarity for contextual words, each context has to be replicated *n*-times. N has to be carefully chosen because each increase in the value of *n* increases the search space exponentially (exponential time complexity $O(V^n)$, v=model vocabulary size). Hence a trade-off between accuracy and computational time has to be carefully resolved. However, preliminary studies revealed that when *n* is set between 50 and 100, optimal performance could be achieved.

Steps used in building the MTM:

- Retrieve the *top-n* most common English words from the BNC based on frequency
- Create contexts by translating each word into its semantic equivalent in other languages using Google translate.
- Replicate each context *n*-times, we used n=100 in our case, this may however depends on the vocabulary size (the number of English words used to create the contexts).
- Train the embeddings with the word2vec CBOW model; we used Gensim a Python library with the following parameter settings: window size (context) =5, negative sampling=5, minimum word count=50, attributes size =300.

## 4.2 Applying the MTM in CLPD

This sub-section describes how we used the MTM model to detect CLP in a suspect document given

I. a collection of source documents in different languages.
II. a source document in a different language (documents in pairs).

### 4.2.1 Using the MTM to detect CLP in a Collection of Source Documents

The application of the MTM in CLPD starts with candidate selection; our approach to candidate selection is similar to that of Ehsan et al., (2016) in that matching words are searched and mapped back to the sentences in which they appear in source and suspect documents using information retrieval, we used the MTM with query expansion and inverted indexing, while Ehsan et al., selected key-words/phrases (1-3 ngrams) in source documents based on *tf-idf* and *tf* scores, and translated the key-words using Google translate. We describe our method in details below.

The MTM is applied in candidate selection through query expansion, query terms are expanded with their translations and used to retrieve potential candidate sources from an inverted index built from a collection of source documents. Query expansion using the MTM involves reformulating a query so that the translations of each query word are retrieved from the MTM. This is possible through vector comparison. The MTM takes a query word and converts it into a word vector and then compares it with word vectors in the model using the cosine measure. The outputs from the vector comparison is a list of words and their similarity to the query word, where the most similar words are the translations of the query word in other languages. For example; when the query word *'friend'* is presented to the MTM, the model outputs;

*'friend'*: [*('ami', 0.99845964), ('freund', 0.99622697), ('amigote', 0.99596620), ('crony', 0.99569368), ('boezemvriend', 0.99561995), ('vriend', 0.99513805), ('amigo', 0.98925573), ('pal', 0.98832619)*]

French=*'ami'*, German=*'freund'*, Spanish=*'amigote'*, *'amigo'*, Dutch=*'vriend'*, *'boezemvriend'*

The output includes synonyms such as *'crony' and 'pal'* in English, *'amigote'* and *'boezemvriend'* in Spanish and Dutch respectively. The model is able to detect related words (synonyms) because semantically similar words do have similar translations, which means they have similar contexts and therefore similar word vectors. When a word is searched, the model outputs the context of the word and similar words in other contexts. This is a key aspect of the MTM, which is one of the advantages it has over online machine translators.

When translations are retrieved for all query words, together they form an expanded query for searching and retrieving potential plagiarised sources from an inverted index. For each candidate document retrieved the original query words (before expansion), and their corresponding matched words in a source document are stored and used for detailed comparison (see figure 1 below) and retrieval of offsets for plagiarised passages.

*Detailed Comparison and Extraction of Plagiarised Passages*
Detailed comparison is carried out on a sentence level, as in previous study (Pataki et al., 2012); this brings the search for plagiarism closer to plagiarised fragments and makes it easier to extract plagiarised passages from clusters of nearby plagiarised sentences. The method used involves mapping the matching word pairs (from the previous stage) to the sentences in which they occur in the source and suspect documents, and normalising the number of matching words by sentence length. Sentences with similarity scores less than a predefined threshold are discarded. When plagiarised sentences are detected, nearby sentences not more than certain characters apart are merged into plagiarised passages; nearby passages are merged, and passages less than certain characters in size are discarded (post-processing).

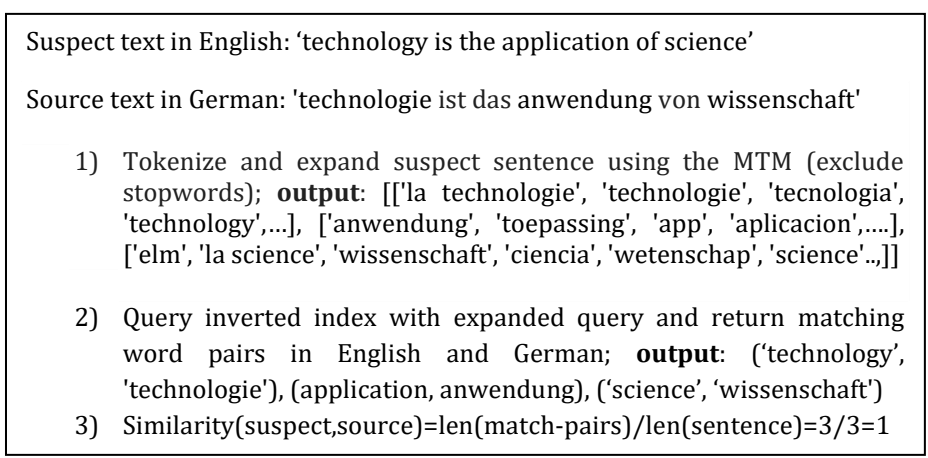

Suspect text in English: 'technology is the application of science'

Source text in German: 'technologie ist das anwendung von wissenschaft'

1) Tokenize and expand suspect sentence using the MTM (exclude stopwords); **output**: [['la technologie', 'technologie', 'tecnologia', 'technology',...], ['anwendung', 'toepassing', 'app', 'aplicacion',....], ['elm', 'la science', 'wissenschaft', 'ciencia', 'wetenschap', 'science'..,]]
2) Query inverted index with expanded query and return matching word pairs in English and German; **output**: ('technology', 'technologie'), (application, anwendung), ('science', 'wissenschaft')
3) Similarity(suspect,source)=len(match-pairs)/len(sentence)=3/3=1

Figure 1: A simple application of the MTM in CLPD

### 4.2.2 Using the MTM Model to Detect Cross-Plagiarism in Document Pairs

When documents/texts are provided in pairs (suspect and source), such as in the SemEval 2016 and 2017 competition, then there is no need for candidate selection, as suspect documents are paired up with their potential sources. The MTM is applied to detect CLP in document pairs as follows:

The method involves comparing sentences in a pair of source and suspect documents using words as units for comparison; it is simply an aggregation of word level similarity to sentence level, and using sentences with similarity scores above a predefined threshold to map out plagiarised passages.

To compute similarity at word level, the MTM is used to compare pairs of words (in source and suspect sentences) by measuring their word vectors using cosine measure, and outputting a similarity score that ranges between *0 and 1*. Word pairs in a pair of sentences with similarity score above a predefined threshold are retained; we used *0.9*7 as our threshold for word pair similarity.

$$sim(word1, word2) = cosine(wordvec1, wordvec2) = \frac{|wordvec1 * wordvec2|}{|wordvec1| * |wordvec2|}$$

To compute similarity at sentence level, the number of similar words in a pair of sentences is normalised by the sentence length (containment). Sentence pairs with similarity scores above a certain threshold are used to map out plagiarised passages in the source and suspect documents by retrieving the offset and length of each sentence, and merging nearby sentences not more than certain distance apart into plagiarised passages. The sentences should line up with the documents they appear in.

## 5   Experiments:

This section describes the evaluation of the proposed CLPD model using datasets that contain cross lingual plagiarism, they include the Pan 2011 and 2012 evaluation corpora on plagiarism detection; both of them are described below.

**The Pan 2011 Evaluation corpus**: this corpus contains 5142 manually and automatic generated cases of CLP distributed across 550 suspect documents of which 4709 were generated automatically using internet translation services, and 433 were generated using both automatic and manual correction processes. The automatic generated cases were created using Google translate to translate text passages from one language to another; the process typically involves removing a passage from a non-English source document, translating the passage into English and inserting it in a document written in English. The manually created cases were artificially generated and corrected by humans to appear like real plagiarism cases. The translations are from {Spanish and Danish) →English.

**The Pan 2012 Evaluation corpus**: this corpus contains artificially generated cases of cross lingual plagiarism distributed across 500 suspect documents. The cross lingual plagiarism cases were created using the multi-lingual Europarl-corpus; from a non-English source document, a text passage is removed and used to retrieve its corresponding English version from the multi-lingual Euro-Parl corpus. The retrieved English version is then inserted into a Gutenberg book (suspicious documents). The translated passages are from {Spanish and Danish} →English.

## 5.1 Evaluation on the Pan 2011 corpus

We started with candidate selection; we built an inverted index on Lucene with the source documents in the collection. We pre-processed each query (suspect document) by case folding and chunking into single words, and extracted key-words using term frequency (TF); we selected words that occur not more than three times in a query. We expanded each query by retrieving translations for words from the MTM, and used the translated words to retrieve relevant source documents from the inverted index table. After retrieving candidate sources for a query document, we carried out detailed comparison between a suspect document and its candidate documents, and mapped out plagiarised passages as described in the method section.

As part of the evaluation, we experimented with MTM models with different vocabulary sizes and measured their performances. This was done in order to determine the best vocabulary size to use for optimal performance of the translation model. The vocabulary sizes experimented with ranges from the top-5000 to the top-45000 most common English words, at an interval of 10000 words. Contexts were created for each word by retrieving the translations of the word (in Danish, French, German, Spanish and Chinese) using Google translate, and each context was replicated 100 times. This experiment was carried out on the Pan2012 corpus because it involves translations between word pairs which can easily be done using the Pan2012 corpus.

Vocab sizes= {5000, 15000, 25000…,.., 45000)

## 5.2 Evaluation on the Pan2012

The Pan@clef 2012 corpus contains documents in pairs; each suspect document is assigned to a source document, comparison is therefore expected to be between pairs of source and suspect documents. Taking this into consideration, we carried out two evaluations with this corpus;

- The first evaluation follows the normal plagiarism detection framework that begins with searching for plagiarised sources in a large collection of source documents; we followed the steps used in the previous evaluation on the Pan2011 corpus.
- The second evaluation involves pairwise document comparison; we used the proposed method for detecting plagiarism in pairs of documents as described in the method section to accomplish this task.

Below are the evaluation metrics and baseline used in this work.

## 5.3 Evaluation metrics

Standard evaluation metrics used in Pan (Potthast et al., 2014; 2015) to measure the performance of plagiarism detection systems were used in this work; they include precision, recall, granularity and pladget score. These measures were applied using positional character alignment (character overlaps) between actual plagiarised passages $S= \{s1,s2....sn\}$ and passages retrieved by a detector

$R=\{r1, r2, ...rn\}$ for a pair of suspect and source documents, and averaged across all plagiarised cases in the corpus.

Precision: measures the proportion of retrieved passages that are relevant

$$prec(S, R) = \frac{1}{R} \sum_{r \in R} \frac{|\bigcup_{s \in S}(s \cap r)|}{|r|}$$

Recall: measures the proportion of relevant passages retrieved.

$$rec(S, R) = \frac{1}{S} \sum_{s \in S} \frac{|\bigcup_{r \in R}(s \cap r)|}{|s|}$$

Granularity: Penalises for multiple (or fragmented) retrieval of a single plagiarism case.

$$gran(S, R) = \frac{1}{|S_R|} \sum_{s \in S_R} |R_S|$$

Pladget score: combines precision, recall and granularity into a single performance score for ranking plagiarism detectors.

$$pladget(S, R) = \frac{F_1}{\log_2(1 + gran(S, R))}$$, F1=harmonic mean of precision and recall.

## 5.4 Baseline:

We re-implemented the Kasprzak and Brandejs (2010) T+MA model as our baseline. We used langdetect; a python language identifier and Google translate to identify languages and translate source documents into the language of suspect documents before applying monolingual plagiarism analysis. For candidate selection, we transformed the source documents in the collection into 5-gram fingerprinting models (using an MD5 hash function) and indexed. To retrieve candidate documents, we transformed a suspect document into a 5-gram fingerprinting model, and used it as query to retrieve all indexed source documents that contain up to 20-matches; gap between any two consecutive chunks should be <=50 chunk length. We then used the matched fragments to retrieve plagiarised passages by retrieving every line of text (offsets) that contains a fragment, and merge neighbouring lines into a passage. See Kasprzak and Brandejs (2010) for the merging parameters.

# 6 Results and Analysis

In this section, the results obtained from the evaluation of the proposed CLPD are presented and analysed. Table 1 contains the results obtained from determination of the most suitable vocabulary size. Tables 2 and 3 contain the results obtained from the evaluation on Pan2011 and 2012 corpora, the baseline and previous studies.

Table 1: Results from the Determination of the Best Vocabulary for an MTM

| Vocabulary size (words) | Precision | Recall | Granularity | Pladget score | Time/sec |
|---|---|---|---|---|---|
| 5000 | 0.875 | 0.802 | 1.020 | 0.819 | 1934.41 |
| 15000 | 0.932 | 0.835 | 1.003 | 0.878 | 2844.79 |
| 25000 | 0.930 | 0.846 | 1.003 | 0.884 | 3068.94 |
| 35000 | 0.926 | 0.850 | 1.003 | 0.884 | 3363.19 |

Table 1 contains the results obtained from the determination of the most suitable vocabulary size for building an MTM with optimal performance.

The main aspect of the results in table 1 is the performance difference observed as the vocabulary size progresses. This reveals the effect of increase in vocabulary size on performance, and the point where the vocabulary size is large enough to build an MTM with optimal performance. The results show that the highest increase in performance was observed between when 5000 and 15000 vocabulary sizes were used (0.819→0.878), and the performance difference decreased significantly and flattens out afterwards. The results suggest that most of the translated CLP cases in the corpus could be detected with an MTM model built with a vocabulary size of the top-15000 most common English words; about 418 out of the 500 cases.

Moving upwards to the top-25,000 most common English words, the models' performance increased slightly, and the run time was still reasonable enough for plagiarism detection tasks; about 6.2 seconds per suspect document. Hence the top-25000 most common English words and their translations in other languages were used in this study as benchmark for building an MTM for CLPD.

In terms of pattern, the results in table 1 show that as the vocabulary size increases, the effectiveness (recall) of the model increases, and the run time increases as well. The recall increases because the models' ability to detect more translated words increases, while the increase in run time is due to increase in search time. The increase in vocabulary size was also met with a corresponding decrease in precision which could be attributed to coincidental matches that occur in the corpus, and may also be due to inaccurate translations of some words. The decrease in precision was however much smaller in comparison to the increase in recall.

Table 2: Evaluation Results on Pan2011 Corpus

| CPLD methods | Precision | Recall | Granularity | Pladget score |
|---|---|---|---|---|
| Manual | | | | |
| Proposed Detector | 0.767 | 0.594 | 1.0065 | 0.667 |
| Previous study | 0.750 | 0.460 | 1.0000 | 0.57 |
| Baseline (T+MA) | 0.727 | 0.445 | 1.0002 | 0.552 |
| Automatic | | | | |
| Proposed detector | 0.954 | 0.943 | 1.0000 | 0.945 |
| Previous study | 0.960 | 0.920 | 1.0000 | 0.94 |
| Baseline (T+MA) | 0.945 | 0.878 | 1.0000 | 0.91 |

Table 2 contains results obtained from the evaluation of the proposed model, the baseline and from a previous study (best performing system in Pan 2011 competition)

The results in table 2 show that the proposed CLPD outperformed (0.667, 0.945) the T+MA baseline (0.552, 0.91) on the manually and automatically generated plagiarism cases. The performance was much higher on the automatic CLP cases (0.945) than on the manually created ones (0.667). Since the manually created cases were simulated to appear like real world cases of CLP, the results therefore point to the difficulty involved in detecting real CLP cases as opposed to machine generated ones, which is consistent with previous studies and the baseline.

In comparison with results obtained from previous studies, the results show that our proposed CLP detector outperformed the best performed system in the Pan 2011 competition; the recall of the proposed CPLD is significantly higher than that of the best performed system, which is a reflection of the effectiveness of the proposed model in terms of being able to detect translated texts that have undergone further alterations, in terms of precision, both systems were more or less even.

Table 3: Results on the Pan 2012 Evaluation Corpus

| CLPD methods | Precision | Recall | Granularity | Pladget score |
|---|---|---|---|---|
| Plagiarism Detection In Document Pairs | | | | |
| Proposed CLPD | 0.93 | 0.846 | 1.003 | 0.884 |
| Baseline (T+MA) | 0.93 | 0.76 | 1.00 | 0.84 |
| Previous study | 0.82 | 0.727 | 1.00 | 0.771 |
| Detection of plagiarism given a source collection | | | | |
| Proposed CLPD | 0.91 | 0.78 | 1.00 | 0.84 |
| Baseline (T+MA) | 0.91 | 0.73 | 1.00 | 0.81 |

Table 3 contains the performance of the proposed CLPD, the baseline and a previous study (best performing system in the Pan2012 competition) on the pairwise CLPD task, and on the standard plagiarism detection task that begins with candidate selection.

The results in table 3 show that the proposed CLPD model outperformed the baseline on the two evaluation tasks undertaken with the Pan 2012 corpus. The difference in performance is seen in the recall (proposed detector: 0.846, 0.78; baseline: 0.76, 0.73), this is due to the fact that the baseline uses 5-gram fingerprinting models for comparison. While this could result in high precision, heavily altered plagiarism cases are unlikely to have significant amount of matching 5-word sequences, but when found, the likelihood of copy is almost certain (see Thompson et al., 2015 for details detailed analysis of how n-gram sizes affect the performance of plagiarism detectors). With respect to the two detection tasks, the performance of the model was much better when plagiarism is detected in pairs of documents, than when plagiarism is detected from a source collection beginning from candidate selection, this pattern is also seen in the baseline and in the previous study as well. The likely reason for the drop in performance is that some of the plagiarised source documents in the corpus may not have been retrieved during candidate selection, and plagiarism cases whose sources were not retrieved were not detected. While the results from the pairwise comparison seem impressive, plagiarism detection is usually preceded with candidate selection. In comparison with previous studies, the results show that the proposed model outperformed the best performed system in the Pan2012 competition on plagiarism detection.

The most likely reason for the superior performance of the proposed CLPD model over the other models is that, the proposed CLPD model is able to detect semantically related words such as synonyms (translated texts that have been paraphrased), and not just words that have been translated directly into their exact meanings in other languages using automatic processes. This is reflected in the significant difference in recall, but not in the precision between both systems. The difference in precision between the proposed CLP detector and the baseline was not significant, which proves that the proposed detector actually captured the translation precision of the T+MA model. The slight difference in precision seen on the Pan2011 corpus (table 2) can be attributed to differences in merging parameters. In addition, the application of sentence level comparison using detected translated texts allows for CLP to be easily detected with much accuracy, which also adds the relatively higher recall of the proposed CLPD model.

# 7 Research Implications

The results presented in this work imply that when embeddings are simulated by projecting English words and their translations (from an online machine translator) in a common embedding space (multilingual contexts), replicated to increase the conditional probability of retrieving semantically similar words, and then trained with the word2vec CBOW model, the trained embeddings capture the translation precision of the online machine translator, and can therefore be used to disambiguate semantically similar texts written in different languages without relying on internet translators. In practical terms, the trained embeddings could be applied in CLPD as an offline machine translator to disambiguate texts written in different languages, and the CLPD model could be effectively used to detect CLP cases that are created automatically by directly translating texts from one language to another, and cases that contain translated texts that have undergone further alterations to evade detection, with state-of-the-art performances. Since the offline translator does not function with internet translators, it could be used for large scale CLPD. The results also imply that, when texts are translated automatically and then paraphrased, they become more difficult to detect than automatically created CLP cases, which is consistent with previous studies. However, CLPD systems that are designed to detect semantically similar words (such as the synonyms of translated words) should be able to detect most difficult cases of CLP. In addition, the results imply that it is possible to detect CLP with state-of-the-art performances without using parallel or comparable corpora, or relying on internet translators.

# 8   Conclusion

In this paper, we addressed the problem of CLP with the aim of proposing a CLP detector that performs as well as a standard T-MA model, but without the limitations faced by the T-MA model. We proposed a model for detecting CLP that uses simulated word embeddings trained using the word2vec CBOW model to capture the translation precision of a T-MA model, and apply them in CLPD without relying on online machine translators. Evaluation against a state-of-the-art T-MA baseline revealed that our proposed CLP detector outperformed the baseline. The findings of this work suggest that a system could be built to detect CLP with precision as high as the T+MA model, and with an overall performance that is even better, but without the limitations. In the future, we will expand the CLP detector to accommodate more languages, mostly non-European languages, and apply the new version in CLPD between languages that have little or no lexical similarity. While in its current form, the proposed CLPD model could be used for monolingual plagiarism detection, it was however evaluated only on CLPD tasks. Future work will therefore include an expansion of the proposed model to include a monolingual component to allow mono and cross lingual plagiarism detection to be carried out.